\documentclass[11pt,a4paper]{article}
\usepackage[hyperref]{acl2021}
\usepackage{times}
\usepackage{latexsym}
\usepackage{multirow}
\usepackage{amsmath}
\usepackage{graphicx}
\usepackage{amsfonts,amssymb}

\usepackage{microtype}
\usepackage{paralist}
\aclfinalcopy 
\def\aclpaperid{2330} 

\title{Bridging Subword Gaps in Pretrain-Finetune Paradigm for \\Natural Language Generation}

\author{Xin Liu\textsuperscript{1,2}~~~Baosong Yang\textsuperscript{3}~~~Dayiheng Liu\textsuperscript{3}~~~Haibo Zhang\textsuperscript{3}~~~Weihua Luo\textsuperscript{3}\\ \textbf{Min Zhang\textsuperscript{4}~~~Haiying Zhang\textsuperscript{1,2}~~~Jinsong Su\textsuperscript{1,2,5\footnotemark[1]}}\\
  \textsuperscript{1}School of Informatics, Xiamen University \\
  \textsuperscript{2}Institute of Artificial Intelligence, Xiamen University\\
  \textsuperscript{3}Alibaba Group~~~
  \textsuperscript{4}Soochow University, China~~~
  \textsuperscript{5}Pengcheng Lab, Shenzhen\\
  
  \texttt{\small{liuxin@stu.xmu.edu.cn}} \\
  \texttt{\small{\{yangbaosong.ybs,liudayiheng.ldyh,zhanhui.zhb,weihua.luowh\}@alibaba-inc.com}}\\
  \texttt{\small{minzhang@suda.edu.cn}}~~~
  \texttt{\small{\{zhang2002,jssu\}@xmu.edu.cn}}

  }

\date{}

\begin{document}
\maketitle

\begin{abstract}
A well-known limitation in pretrain-finetune paradigm lies in its inflexibility caused by the one-size-fits-all vocabulary.  
This potentially weakens the effect when applying pretrained models into natural language generation (NLG) tasks, especially for the subword distributions between upstream and downstream tasks with significant discrepancy. 
Towards approaching this problem, we extend the vanilla pretrain-finetune pipeline with an extra embedding transfer step. 
Specifically, 
a plug-and-play embedding generator is introduced to produce the representation of any input token, according to pre-trained embeddings of its morphologically similar ones. 
Thus, embeddings of mismatch tokens in downstream tasks can also be efficiently initialized. 
We conduct experiments on a variety of NLG tasks under the pretrain-finetune fashion. 
Experimental results and extensive analyses show that the proposed strategy offers us opportunities to feel free to transfer the vocabulary, leading to more efficient and better performed downstream NLG models.
\footnote{We release the code at https://github.com/ DeepLearnXMU/embedding-transfer}


\end{abstract}

\renewcommand{\thefootnote}{\fnsymbol{footnote}}
\footnotetext[1]{Jinsong Su is the corresponding author. This work was done when Xin Liu was interning at DAMO Academy, Alibaba Group.}

\section{Introduction}

Pretrain-finetune paradigm has been highly successful on tackling challenging problems in natural language processing, e.g., domain adaptation~\cite{DBLP:conf/emnlp/SatoS0TK20,coling/YaoYZCL20}, incremental learning~\cite{DBLP:conf/aclnmt/KhayrallahTDK18,DBLP:conf/emnlp/WanYWZCZC20}, as well as knowledge transferring~\cite{DBLP:conf/ijcai/LiuLLSGWL20}.  
The rise of large-scale pre-trained language models further attracts increasing attention towards this strategy~\cite{DBLP:conf/naacl/DevlinCLT19, DBLP:conf/naacl/EdunovBA19}.
Typically, these methods first pretrain a universal model using a large-scale corpus, which is then finetuned to various downstream tasks via a few adjustments.
Due to its simplicity yet impressive performance, pretrain-finetune paradigm becomes the undoubtedly dominant solution for building state-of-the-art models in many natural language understanding tasks~\cite{DBLP:conf/naacl/XuLSY19,DBLP:conf/acl/YangWLLLWSL19,DBLP:conf/ijcai/LiuLLSGWL20}.



\begin{table}[t]
\centering
\resizebox{0.49\textwidth}{!}{
\begin{tabular}{ll}
\hline
\hline
\multicolumn{1}{l|}{M-BERT} & Ce\textcolor{red}{\_}no\textcolor{red}{\_}zo\textcolor{red}{\_}ic  pala\textcolor{red}{\_}eo\textcolor{red}{\_}hy\textcolor{red}{\_}dro\textcolor{red}{\_}dyn\textcolor{red}{\_}ami\textcolor{red}{\_}c  \\
\multicolumn{1}{l|}{Out-of-Domain}    & Cen\textcolor{red}{\_}ozo\textcolor{red}{\_}ic  pal\textcolor{red}{\_}a\textcolor{red}{\_}e\textcolor{red}{\_}o\textcolor{red}{\_}hydro\textcolor{red}{\_}dynamic \\ \hline
\multicolumn{1}{l|}{Thesis}   & Cenozoic palaeohydrodynamic  \\
\hline\hline
\end{tabular}}
\caption{Segmentation of English sequence ``\emph{Cenozoic palaeohydrodynamic}'' learned from different data distribution as described in \S~\ref{sec:exp}. High frequent words in thesis domain are split into fine-grained and under-represented tokens in pre-trained models.}
\label{segment_example}
\end{table}
In comparison, this strategy often achieves disappointing or barely satisfactory performance in natural language generation (NLG) tasks. For example, several studies observe that M-BERT \cite{DBLP:conf/naacl/DevlinCLT19} fails to enhance the decoder of a translation model~\cite{DBLP:conf/naacl/EdunovBA19,DBLP:conf/iclr/ZhuXWHQZLL20}, while \newcite{DBLP:journals/tacl/RotheNS20} reach the same conclusion even when adapting an autoregressive model GPT~\cite{radford2019language}. 
A natural problem arises: \textit{What is the crucial bottleneck in current pretrain-finetune framework and how to break it?}



In this paper, we provide the first answer from the subword discrepancy aspect, namely, the subword vocabulary extracted according to the pre-training data distribution is insufficient to cope with the downstream NLG tasks. 
Such inflexibility stems from the fact that downstream NLG models have to inherit the vocabulary from their pre-trained counterparts.
In order to deal with the open-vocabulary problem, it is de-facto standard for pre-trained models to employ 
heuristic subword segmentation methods~\cite{DBLP:conf/acl/SennrichHB16a,DBLP:conf/emnlp/KudoR18}. 
However, the segmentation learns on the upstream corpus other than the finetuned data and is likely to be sub-optimal~\cite{DBLP:conf/emnlp/CherryFBFM18,DBLP:conf/acl/ProvilkovEV20}. 

We argue that these lead to subword discrepancy and bring two defects.  
Firstly, the pre-trained model usually learns a fine-grained subword segmentation to maintain the coverage of a large amount of diverse vocabulary. Consequently, downstream NLG models may suffer from more serious exposure bias~\cite{DBLP:conf/nips/BengioVJS15} and expensive computational cost caused by the increased sequence lengths. As one example, M-BERT exploits 100 thousand fine-grained subwords to encode hundreds of languages, while most of downstream NLG tasks, in fact, require only one language and its associate tokens. 
Secondly, words that are rare in upstream task but frequent in downstream task may be segmented end up poorly understood~\cite{DBLP:conf/acl/ProvilkovEV20}.
Considering the English sequence ``\emph{Cenozoic palaeohydrodynamic}'' shown in Table~\ref{segment_example}, all the words are frequent in a thesis domain translation task and can be well preserved in its vocabulary. Nevertheless, they are segmented into under-represented tokens by pre-trained models, preventing the finetuning stage from better learning their compositionality for generation. 
An alternative solution is reconstructing the pre-trained model by exploiting either a task-specific vocabulary \cite{DBLP:conf/ijcnlp/NguyenC17, DBLP:conf/wmt/KocmiB18} or a subword regularization approach \cite{DBLP:conf/acl/ProvilkovEV20}. However, retraining the upstream model from scratch for each task is time-consuming and unavailable for large-scale models like M-BERT, GPT, etc. 

\vspace{-0.2em}
To this end, we propose a simple yet generalized pretrain-finetune strategy, where an embedding transfer stage is inserted between pre-training and finetuning to eliminate their token granularity gaps.  
Unlike the prior strategy using a fixed vocabulary, our vocabulary is changeable and its items including mismatched ones can be easily initialized by the pre-trained embeddings. Concretely, we equip the pre-trained model with a plug-and-play embedding generator, which is able to produce the embedding of any token by feeding its subwords and hyperwords that appeared in pre-trained vocabulary.  
To train this generator, we randomly split or merge some tokens to replace their original embeddings with those produced by the generator. The parameters of the generator are optimized under the vanilla pre-training framework to minimize the divergence before and after replacing the embeddings. 
Accordingly, we can use a task-specific vocabulary for the downstream task,
where common tokens are immediately initialized with pre-trained embeddings while mismatched ones are initialized by our generator.

We conduct experiments on various tasks under NLG context, in a range from domain adaptation to knowledge transferring, and from machine translation to answer-aware question generation. Empirical results demonstrate the universal-effectiveness of the proposed strategy comparing with strong baselines and related approaches. Quantitative and qualitative analyses verify that tackling subword discrepancy can exactly alleviate the problem of exposure bias, large computational cost, and the under-represented tokens in vanilla pretrain-finetune paradigm. 
To summarize, the contributions of our work are as follows:
\begin{compactitem}
\item Through in-depth analyses,
we point out and formally analyze 
subword discrepancy,
affecting the conventional pretrain-finetune strategy in NLG tasks.
\item We propose a simple, flexible, and generalized pretrain-finetune training strategy, where an embedding generator is introduced to leverage the knowledge of the pre-trained model to initialize embeddings of any required tokens. 
\item Extensive experiments show that our strategy is able to efficiently decrease the vocabulary gaps in pretrain-finetune paradigm and significantly boost the performance of NLG models.
\end{compactitem}

\section{Related Work} 
Recent studies observe that pre-trained models suffer a bottleneck when they are applied to NLG tasks \cite{DBLP:conf/naacl/EdunovBA19, DBLP:conf/iclr/ZhuXWHQZLL20, DBLP:journals/tacl/RotheNS20}. 
This problem has been attributed to many reasons. For example,~\newcite{DBLP:conf/nips/YangDYCSL19} point out pretrain-finetune discrepancy caused by the absent masked frames in real data when adopting pre-trained masked language models. \newcite{DBLP:conf/naacl/ChronopoulouBP19} investigate catastrophic forgetting in finetuning stage. It can be said that how to successfully employ pretrain-finetune to enhance NLG models remains a great challenge. 
We explore this problem from another direction, i.e., the unsuitable subword segmentation for downstream tasks. 


\paragraph{Task-Specific Vocabulary} 
A natural manner to address this issue is to adopt a task-specific vocabulary. 
\citet{DBLP:conf/acl/LewisLGGMLSZ20} first replace the embedding layer with an independent encoder, of which vocabulary and parameters are learned from the downstream corpus. Along this line, \citet{DBLP:conf/emnlp/SatoS0TK20} exploit external monolingual data to construct a new embedding layer and achieve improvements in domain adaptation. 
This series of studies empirically confirm the necessity of the suitable vocabulary for the finetuning stage. However, these methods have to learn the task-specific embeddings separately before each adaptation, which brings in additional computational cost thus limiting their applicability. Besides, they completely discard the pre-trained embeddings, which have been proved to be useful by \newcite{DBLP:conf/acl/AjiBHS20}. Extra encoder or embedding layer may fail to be well optimized with insufficient downstream resources. 
Accordingly, \citet{DBLP:journals/tacl/RotheNS20} employ a task-specific vocabulary to retrain M-BERT, which is then used to initialize neural machine translation (NMT) model. Considering more robust approaches, \newcite{DBLP:conf/acl/Kudo18} and \newcite{DBLP:conf/acl/ProvilkovEV20} randomly sample segmentations for each sentence at the training time. 
Unlike the above methods, our goal is to build a plug-and-play component, that involves neither retraining the pre-trained model nor learning task-specific embeddings separately.

\paragraph{Embedding Generator} Our work is also related to studies with respect to generating embeddings for out-of-vocabulary (OOV) words.
In this context, researchers use embeddings of characters or subwords to predict those of unseen words \cite{DBLP:conf/emnlp/PinterGE17, DBLP:conf/emnlp/ZhaoML18, DBLP:conf/naacl/SasakiSI19, DBLP:conf/emnlp/Fukuda0K20}. For example, \citet{DBLP:conf/emnlp/ZhaoML18} train an embedding generator through reconstructing the original representation of each word from its bag of subwords. \newcite{DBLP:conf/naacl/SasakiSI19} progressively improve the generator using attention mechanism. \newcite{DBLP:conf/emnlp/Fukuda0K20} further leverage similar words to enhance this procedure. 
Our work significantly differs from the above studies in two aspects.
Due to the vocabulary is fixed once predefined, the embedding reconstruction can be merely drawn on a few of selected words.
By contrast, our generator is able to produce embeddings of any tokens, since these embeddings are directly embedded into the pre-trained model with an objective in terms of minimizing the divergence. 
Moreover, previous studies mainly focus on handling the problem of OOV, while our work, to our best of knowledge, is the first study that exploits embedding generator to transfer granularity over subwords for pretrain-finetune paradigm.

\section{Methodology}
\label{sec:method}

In this section,
we introduce our proposed pretrain-finetune strategy in detail.

\subsection{Main Steps in Our Strategy}

\begin{figure}[h] 
\centering 
\includegraphics[width=0.45\textwidth]{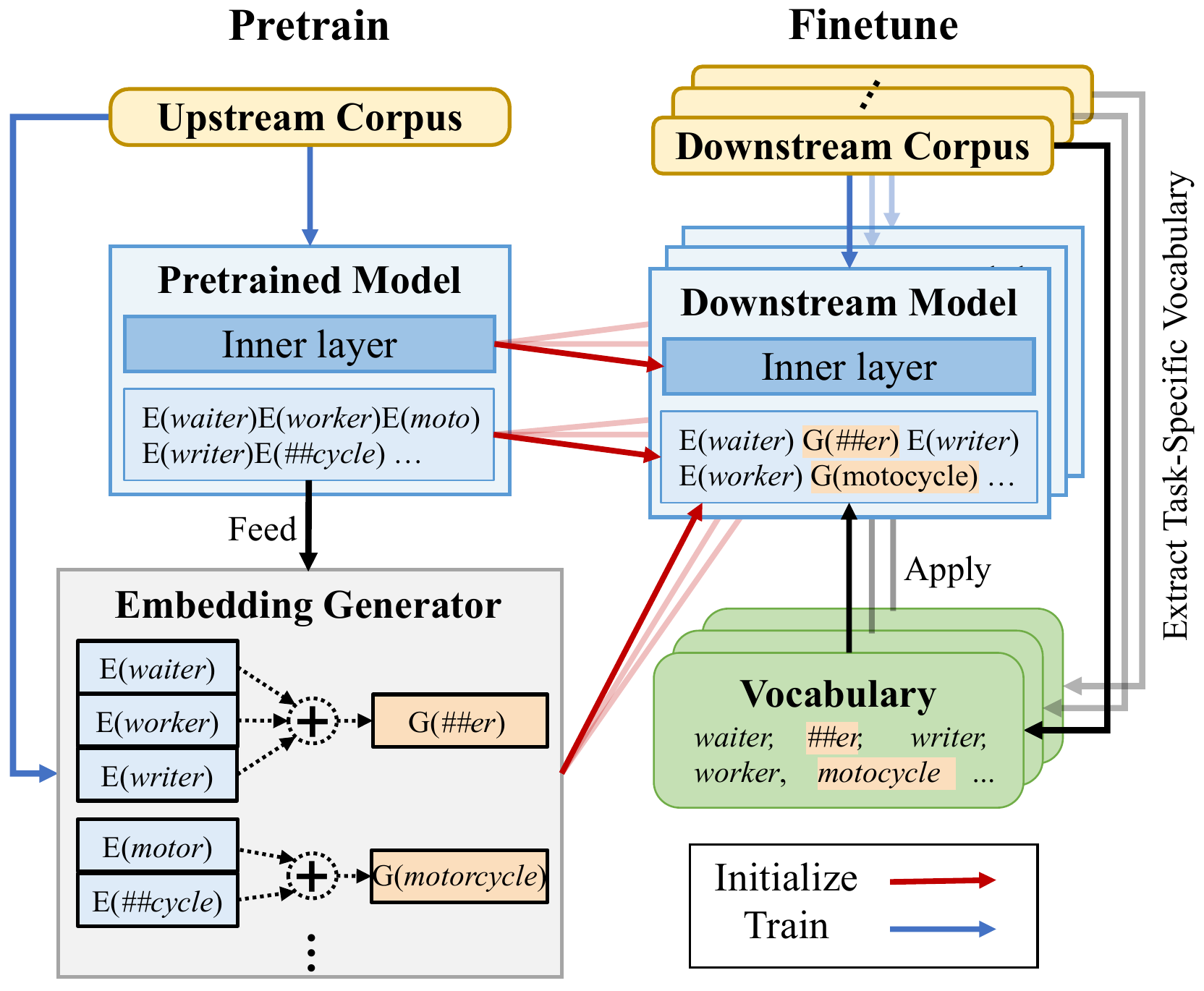} 
\caption{Illustration of our pretrain-finetune pipeline. We pretrain an embedding generator for the initialization of embeddings of unseen tokens. Thus, each downstream model can adopt its suitable vocabulary instead of the unchangeable one. $E(\cdot)$ and $G(\cdot)$ indicate the pretrained and generated embedding, respectively. } 
\label{pipeline}
\end{figure}

As shown in Figure \ref{pipeline}, we extend the prior pretrain-finetune paradigm with an embedding transfer stage. 
Specifically, we revise the conventional pretrain-finetune pipeline as follows:

\noindent\textbf{Pretrain.} As usual, we first construct a pre-trained model using an existing large-scale corpus. In addition, we further pretrain an embedding generator regardless of downstream tasks. It's expected to produce the embedding of any required token, by feeding pre-trained embeddings of its subwords and hyperwords. Hence, it can be employed into any downstream tasks for embedding transferring.  

\noindent\textbf{Finetune.} We differently initialize the word embeddings and the other parameters (inner layer) for the downstream model, respectively. For the former, we use the downstream-task training corpus to learn a task-specific subword segmentation and corresponding vocabulary. 
For an unseen token, we apply the generator to produce its initial representation. Otherwise, we directly initialize it with the corresponding pre-trained embeddings. Considering the latter, we directly adapt inner-layer parameters of the pre-trained model to the downstream model. Finally, we continue to train the downstream model using the finetuning data following the common fashion. 

As seen, our strategy is lightweight and also able to avoid the issue of subword discrepancy, since it does not require retraining for the pre-trained model and can be quickly applied to various downstream NLG models. 
\subsection{Constructing the Embedding Generator}
To make the word embedding generator applicable to all downstream NLG models, we design the generator so that it can generate the embedding of any input token according to those of its morphologically similar tokens from the learned pre-training vocabulary.
The basic intuition behind our design stems from this fact:
if the input token is a complete word, like \emph{motorcycle}, 
its semantic meaning is related to those of its \textbf{subword}s, \emph{motor} and \emph{\#\#cycle}.
On the contrary, 
if the input token is a subword, such as \emph{\#\#er}, the words that contain the input token, which we call them \textbf{hyperword}s,
e.g., \emph{worker}, \emph{writer} and \emph{singer},
can be exploited to learn its semantic meaning.

Concretely,
given a mismatch token $w$,
we borrow the segmentation principle from pre-trained model to split $w$ into subwords based on the pre-training vocabulary,
and traverse the pre-training vocabulary to select all longer tokens containing $w$.
Then,
we combine the generated subwords and the selected hyperwords to form the morphologically similar token set of $w$, denoted by $S_m(w)$. 
Afterwards, 
we explore three kinds of generators to produce the embedding $G(w)$ of $w$:

\paragraph{AVG-EG: Averaging-Based Embedding Generator}
Intuitively, we can simply define $G(w)$ as the average embedding of the words from $S_m(w)$:
\begin{equation}
G(w)=\frac{1}{|S_m(w)|}\sum_{w'\in S_m(w)}E(w')\text{,}
\end{equation}
where $E(w')$ is the pre-trained embedding of the token $w'$. 
In this way,
our generator can be directly used, without increasing the cost of training time.

\paragraph{ATT-EG: Attention-Based Embedding Generator}
Another natural solution is to softly fuse information from different morphologically similar words using an attention mechanism~\cite{DBLP:journals/corr/BahdanauCB14}. The $G(w)$ is formally expressed as:
\begin{equation}
    \begin{aligned}
    G(w)&=\frac{1}{|S_m(w)|}\sum_{w'\in S_m(w)}\alpha(w')\cdot E(w')\text{,}\\
    \alpha(w')&=\frac{\text{exp}(\mathbf{W}^\top E(w'))}{\sum_{w''\in S_m(w)} \text{exp}(\mathbf{W}^\top E(w''))}\text{,}
    \end{aligned}
\end{equation}
where $\mathbf{W}\in \mathbb{R}^{1\times d}$ indicates a learnable vector, $d$ denotes the dimensionality of word embedding. 
Compared with the first generator,
this generator can be jointly trained with the pre-trained model,
therefore it is capable of better quantifying the effects of morphologically similar words in $S_m(w)$.
\begin{figure*}[h] 
\centering 
\includegraphics[width=1.0\textwidth]{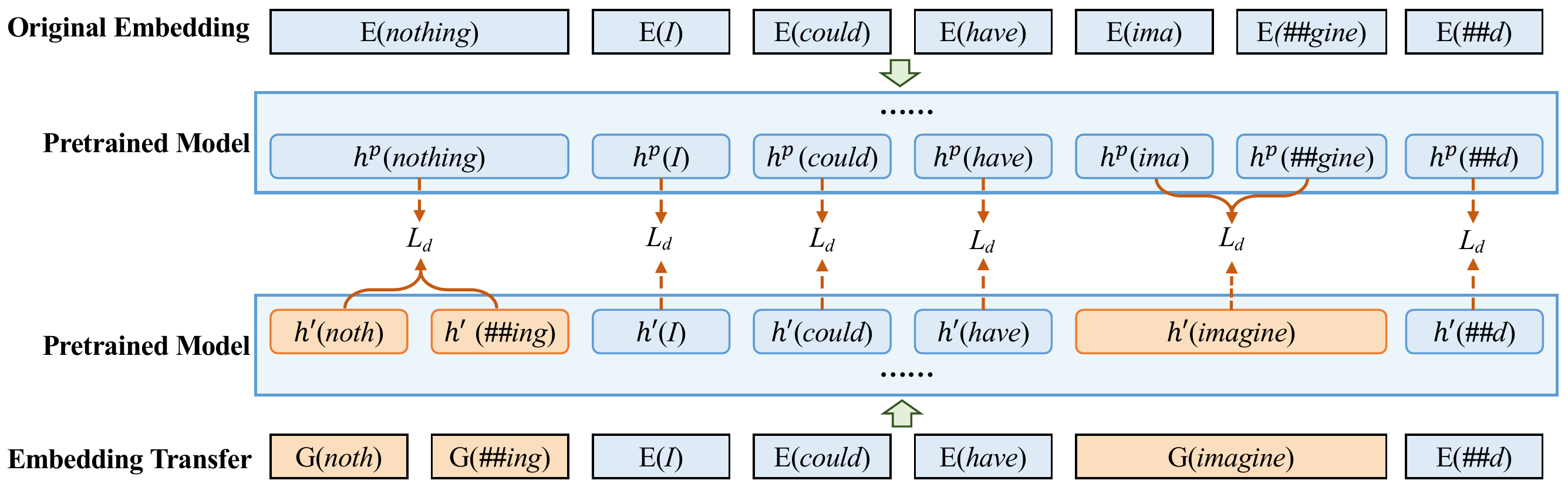} 
\caption{Illustration of the knowledge distillation procedure. Our strategy first performs a segmentation (differs from the pre-trained one) on the original sentence to create unseen tokens, of which embeddings can be produced by our embedding generator. We fix the inner layers of the pre-trained model and force our model to 
narrow the distance between its output layer and the conventional one. } 
\label{s&m}
\end{figure*}

\paragraph{PATT-EG: Position-Aware Attention-Based Embedding Generator}
From the linguistic perspective, different locations of morphemes in a word reflect distinct semantic meaning. Consequently, we refine the above attention-based generator by considering six kinds of morphology relationships between $w$ and $w'\in S_m(w)$: if $w'$ is a subword of $w$, $w'$ can be the prefix/infix/suffix subword of $w$. In turn, if $w'$ is a hyperword of $w$, $w$ can be the prefix/infix/suffix subword of $w'$.
Formally,
$G(w)$ is produced in the following way:
\begin{equation}
\begin{aligned}
G(w)&=\frac{1}{|S_m(w)|}\sum_{w'\in S_m(w)}\alpha(w') E(w')\text{,}\\
\alpha(w')&=\frac{\text{exp}(\mathbf{I}\mathbf{W_r} E(w'))}{\sum_{w''\in S_m(w)} \text{exp}(\mathbf{I}\mathbf{W_r} E(w''))}\text{,}
\end{aligned}
\end{equation}
where $\mathbf{W_r}\in \mathbb{R}^{6\times d}$ is a learnable parameter matrix, and $\mathbf{I}\in \mathbb{R}^{1\times6}$ is the one-hot vector indicating the relationship between $w$ and $w'$.

Note that, all the trainable generators are designed to lightweight architectures with a few of parameters. We believe this can achieve a more generalizable model and speed up their convergence. We will compare and investigate these generators in the subsequent experiment section.
\subsection{Training the Embedding Generator} \label{sec:training} 
One principle of our strategy is plug-and-play, 
which can be directly applied to initialize any unseen tokens in all downstream NLG tasks, avoiding the time cost of retraining the model. 
To this end, we borrow the pre-trained model and its associated corpus to train our generator before finetuning.

In the specific implementation,
we first preprocess the sentences of pre-training corpus,
where two kinds of preprocessing operations are applied to simulate unseen tokens: 1) randomly selecting some consecutive subwords and combining them into an unseen token; and 2) randomly choosing a token and splitting it into several consecutive unseen tokens. 
Figure \ref{s&m} provides an example of sentence preprocessing, where the word \emph{nothing} is randomly split into two unseen subwords \emph{noth} and \emph{\#\#ing},
while the subwords \emph{ima} and \emph{\#\#gine} are concatenated into an unseen token \emph{imagine}. 
Through this data preprocessing,
we can obtain large amounts of samples with unseen tokens involving various granularities,
which facilitates the robustness of our generator. 

Then, 
we embed our generator into the pre-trained model to encode unseen words, and \textbf{fix} parameters of the pre-trained model to train the generator according to the following objectives:
\paragraph{Reusing Pre-training Loss}
The generated embeddings should share the same latent space with the existing embeddings, in the meanwhile, representing appropriate semantic meaning. Accordingly, we serve to minimize the vanilla loss of pre-trained model as the basic training objective of our generator. The loss function can be diverse according to the upstream tasks, which is denoted as $\mathcal{L}_p(s')$ with $s'$ being the preprocessed training sentence. 
\paragraph{Knowledge Distillation} We further exploit knowledge distillation~\cite{DBLP:journals/corr/HintonVD15} to narrow the divergence between hidden states in the pre-trained model before and after applying the generated embeddings. Given a training example $s$, the vanilla pre-trained model and our generator preprocess it to $s^p$ and $s'$, respectively. As shown in Figure \ref{s&m}, we transfer the knowledge of the output layer in terms of $s^p$ to that of $s'$. Euclidean Distance is adopted to measure the divergence between representations output by vanilla pretrained model $h^p(w)$ and that of our model $h'(w)$ with respect to the same word $w$. Since each word may be split into different sequences of tokens, we regard the average hidden states of the corresponding token sequence as its representation. Thus, the loss function can be defined as:
\begin{equation}
\begin{aligned}
\mathcal{L}_{d}(s^p,s')=\frac{1}{|s|}\sum_{w \in s}
||h^p(w)-h'(w)||^2 \text{,}
\end{aligned}
\end{equation}

Finally, we assign a hyper-parameter $\lambda$ to quantify the effect of $\mathcal{L}(\cdot)$ and $\mathcal{L}_{d}(\cdot)$, which is empirically set to 0.5 as default:
\begin{equation}
\mathcal{L}(s^p,s') = \mathcal{L}_p(s') + \lambda \mathcal{L}_{d}(s^p,s') \text{.}
\end{equation}

\section{Experiments} \label{sec:exp}
In this section, we examine the effectiveness of the proposed strategy in a variety of NLG tasks. We first run a set of experiments to compare the variants of our approach and the related methods on domain adaptation translation tasks. Then, we assess the superiority of our approach on transferring the knowledge from M-BERT~\cite{DBLP:conf/naacl/DevlinCLT19} and M-BART~\cite{liu2020multilingual} to two downstream NLG tasks: machine translation (MT) and answer-aware question generation (QG). 

\subsection{Domain Adaptation}
We conduct experiments on English-to-Chinese (En$\Rightarrow$Zh) domain adaptation translation tasks, where the pretrain-finetune paradigm resort as standard. The pre-training corpus is extracted from an out-of-domain dataset LDC\footnote{Including LDC2002E18, LDC2003E07, LDC2003E14, LDC2004T07, LDC2004T08 and LDC2005T06.}, in which 1.25M (M = million), 3K (K = thousand), 3K sentences pairs are randomly sampled as training, development and test set, respectively. We verify the effectiveness of our strategy on two downstream domains: Thesis and Laws, of which data are collected from UM-Corpus \cite{DBLP:conf/lrec/TianWCQOY14}. 
We follow the same settings as \newcite{DBLP:conf/emnlp/ZengSWLXYZ18} and \newcite{jiali_tpami} to preprocess two corpus and train models. The translation quality is evaluated by cased BLEU \cite{DBLP:conf/acl/PapineniRWZ02}, which is caculated by \emph{mteval-v13a.pl}.




\paragraph{Implementation Details}
All the compared methods are re-implemented on top of \textit{FairSeq}\footnote{\url{https://github.com/pytorch/fairseq}} and built on Transformer \cite{DBLP:conf/nips/VaswaniSPUJGKP17}. We apply Adam Optimizer \cite{DBLP:journals/corr/KingmaB14} with $\beta_1$ and $\beta_2$ being 0.9 and 0.999, respectively. The dropout ratio is set to 0.3 and each iteration batch consists of 25K tokens. 
For both pre-training and finetuning, we employ warm-up strategy where the linear warm-up phase takes 4K steps, reaching its maximum learning rate to $5\times10^{-4}$.
The training of each model is early-stopped to maximize BLEU score on the development set.
Other hyperparameters are set following \emph{Base} setting in \newcite{DBLP:conf/nips/VaswaniSPUJGKP17}. We investigate the following methods: \footnote{Hyperparameters that are not mentioned in our paper are set to the default according to the corresponding literatures.}
 \begin{compactitem}
    \item \emph{Baseline}: We design baselines under two basic settings: \textbf{Single-Run} denotes that the translation model only trained on in-domain corpus with the domain-specific vocabulary. \textbf{Pretrain-Finetune} represents the well-known pipeline, i.e., pre-training using upstream corpus, then finetuning on in-domain dataset via inheriting pre-training vocabulary.
    \item \emph{Task-Specific Vocabulary}: This group of methods retrain the upstream model using a task-specific vocabulary, involving: the vocabulary collected from in-domain data~\citep[\textbf{Downstream Vocab},][]{DBLP:journals/tacl/RotheNS20}, the joint vocabulary extracted from all corpus~\citep[\textbf{Joint Vocab},][]{DBLP:conf/ijcnlp/NguyenC17}, as well as the pre-trained vocabulary with a subword regularization process on upstream corpus for robustness~\citep[\textbf{BPE-Drop},][]{DBLP:conf/acl/ProvilkovEV20}.
    \item \emph{Embedding Generator}: We also examine several representatives of existing embedding generators on pretrain-finetune paradigm. 
    We assign the domain-specific vocabulary for each downstream model, in which embeddings of the seen tokens are reused, while the mismatched ones are: 1) randomly initialized~\citep[\textbf{Random Init},][]{DBLP:conf/acl/AjiBHS20}; 2) learned by \textbf{Word2Vec} \cite{DBLP:journals/corr/abs-1301-3781} using in-domain data; and 3) produced by a generator trained via reconstructing embeddings using Bag-of-Subwords~\citep[\textbf{Embedding Recon},][]{DBLP:conf/emnlp/ZhaoML18}. 
    \item \emph{New Embedding Layer}: These methods assigned the domain-specific vocabulary for each downstream model, but completely discard the embeddings of upstream models. The new embeddings are produced from: 1) randomly initialized \textbf{Independent Encoder} \cite{DBLP:conf/acl/LewisLGGMLSZ20}; and 2) \textbf{CBOW} model trained under the downstream corpus \cite{DBLP:conf/emnlp/SatoS0TK20}.
    \item \emph{Our Strategy}: Our embedding generators are trained using the setting of pre-trained model with one epoch, as described in \S~\ref{sec:method}.
\end{compactitem}
Note that, to eliminate the influence of control variables, all the vocabulary transfers in above models are conducted on the decoder-side only. 



\begin{table}[]
\centering
\scalebox{0.9}{
\begin{tabular}{llcc}
\hline \hline
\multicolumn{2}{c|}{Strategy}                                                & Thesis               & Laws \\ \hline
\multicolumn{4}{c}{\emph{Baseline}}                                                                             \\ \hline
\multicolumn{2}{l|}{\   Single-Run}                                    & 34.51                & 52.21    \\
\multicolumn{2}{l|}{\   Pretrain-Finetune}                        & 30.21                & 52.12    \\\hline
\multicolumn{4}{c}{\emph{Task-Specific Vocabulary}}                                                                             \\ \hline
\multicolumn{2}{l|}{\   Downstream Vocab}                          &  31.70 & 52.23     \\ 
\multicolumn{2}{l|}{\   Joint Vocab}    & 35.01                & 52.70    \\
\multicolumn{2}{l|}{\   BPE-Drop}    & 32.41                & 52.43    \\
\hline
\multicolumn{4}{c}{\emph{Embedding Generator}}                                                                      \\ \hline
\multicolumn{2}{l|}{\  Random Init}     & 36.33                & 53.14    \\    
\multicolumn{2}{l|}{\  Word2Vec} & 36.21                & 53.11    \\ 
\multicolumn{2}{l|}{\  Embedding Recon}  & 36.25                   & 53.01    \\ \hline
\multicolumn{4}{c}{\emph{New Embedding Layer}}                                                                      \\ \hline
\multicolumn{2}{l|}{\  Independent Encoder}     & 34.77                & 52.73    \\    
\multicolumn{2}{l|}{\  CBOW} & 36.12                & 52.93    \\ \hline
\multicolumn{4}{c}{\emph{Our Strategy}}                                                                            \\ \hline
\multicolumn{2}{l|}{\  AVG-EG}                                   & 37.03                & 53.30    \\
\multicolumn{2}{l|}{\  ATT-EG}                                   & 37.40                & 53.39    \\
\multicolumn{2}{l|}{\  ~~~+Knowledge   Distillation}                                  & 37.59                    & 53.87    \\ 
\multicolumn{2}{l|}{\  PATT-EG}                                  & 37.72                    & 53.85    \\ 
\multicolumn{2}{l|}{\  ~~~+Knowledge   Distillation}                                  & \textbf{37.90}                    & \textbf{54.27}    \\ 
\hline \hline
\end{tabular}
}
\caption{Evaluation (BLEU) of different pretrain-finetune strategies on En$\Rightarrow$Zh domain translation tasks. }
\label{domain_adapt}
\end{table}

\begin{table*}[]
\centering
\scalebox{0.9}{
\begin{tabular}{l|rrr|rrr|rrlll}
\hline \hline
\multirow{2}{*}{\bf Models}  & \multicolumn{3}{|c}{\bf WMT14 En$\Rightarrow$De}  &  \multicolumn{5}{|c}{\bf SQuAD v1.1 Question Generation}\\
    \cline{2-9}
        &   BLEU    &   \# Param.    &  Speed & ROUGE-L & BLEU & METEOR &  \# Param.  & Speed \\ \hline
            \hline 
Random Init & 26.08 & 382M & 25.64  & 23.98 & ~2.91 & 9.25 & 382M & 21.23 \\ 
\hline 
w/ M-BERT  & 28.24 & 382M & 26.51  & 25.88 & ~3.31 & ~9.27 &382M & 21.58 \\ 
~~~+Ours  & 29.77 & 255M & 49.54  & 26.76 & ~3.55 & ~9.86 & 242M & 27.86 \\ \hline
w/ M-BART & 29.13 & 610M & 19.65  & 48.07 & 20.20 & 24.29 & 610M & 12.62 \\
~~~+Ours & \textbf{30.15} & 387M & 25.79  & \textbf{48.11} & \textbf{20.27} & \textbf{24.31} & 363M & 14.41 \\ \hline \hline

\end{tabular}
}
\caption{Evaluation of our model on knowledge transferring tasks. \emph{``w/''} denotes ``with''. Random Init uses the same architecture as `` w/ M-BERT'' while being initialized randomly. ``\# Param." denotes the trainable parameter size of each model. ``Speed'' indicates the inference speed measured in sentences per second.\protect\footnotemark} 
\label{table_mt}
\end{table*}

\paragraph{Results}
Table \ref{domain_adapt} lists our results on domain adaptation tasks. Considering baseline models, immediately finetuning a downstream model with out-of-domain vocabulary performs worse than merely training each model using in-domain data and task-specific vocabulary. This is consistent with findings in~\newcite{DBLP:conf/naacl/EdunovBA19} and ~\newcite{DBLP:conf/iclr/ZhuXWHQZLL20}. We observe that there are over 13K and 11K tokens in the vocabulary in terms of Out-of-Domain are mismatched with that of Thesis and Laws respectively, indicating that subword discrepancy indeed harms the performance of downstream NLG models. When adapting task-specific vocabulary to retrain upstream models, all the translation qualities are improved, 
confirming the necessity of bridging subword gaps between upstream and downstream models.  
In addition, we also appraise several existing embedding transfer strategies into pretrain-finetune pipeline. Interestingly, randomly initializing embeddings of unseen tokens yields even slightly better results than utilizing ``Word2Vec'' and ``Embedding Recon''. We attribute this to the fact that the training of the latter two generators is individual regardless of the pre-trained model, resulting in unshared latent space between the generated and pre-trained embeddings.   

Our models surpass all baselines and related methods on translation qualities. Most importantly, in contrast to existing approaches that have to either retrain the pre-trained model from scratch or learn a separate embedding generator for each domain, our strategy can be immediately adopted to any downstream tasks once ready. Specifically, PATT-EG achieves the best performance, confirming our hypothesis that softly summarizing information from morphologically similar tokens and considering positions of morphemes facilitate the embedding transferring. Besides, using knowledge distillation to narrow the divergence before and after applying our generator can progressively improve the performance. Accordingly, we use \textbf{PATT-EG + Knowledge Distillation} as the default setting in subsequent experiments.


\footnotetext{Single NVIDIA v100 GPU with batch size being 32.}

\subsection{Knowledge Transferring}
We test our method on transferring the knowledge from two advanced large-scale language models: non-autoregressive M-BERT and autoregressive M-BART. For computational efficiency, we randomly extract 4M samples from the conventional pre-training corpus\footnote{\url{https://dumps.wikimedia.org}} to train our embedding generator using the configurations of pre-trained models with one epoch and 4,096 batch size. Comparisons are conducted on machine translation and question generation task. The pre-trained model is employed on both of encoder and decoder. Same as configurations in domain adaptation, we merely perform the embedding transferring in decoder. Since the two language models exploit different segmentation tools, i.e., WordPiece~\cite{DBLP:journals/corr/WuSCLNMKCGMKSJL16} and SentencePiece~\cite{DBLP:conf/acl/Kudo18}, we set 32K and 10K as the number of word and sentence pieces for downstream tasks, respectively.
\vspace{-0.3em}
\paragraph{Machine Translation}
Considering machine translation, we examine our method on the widely used English-to-German (En$\Rightarrow$De) benchmarks: WMT14. We follow \newcite{DBLP:journals/tacl/RotheNS20} and \newcite{liu2020multilingual} to deal this task.
\vspace{-0.3em}
\paragraph{Question Generation} We use the SQuAD v1.1 \cite{DBLP:conf/emnlp/RajpurkarZLL16} dataset for question generation.
We follow the common setting to preprocess dataset and train our models \cite{DBLP:journals/corr/abs-2011-11928}.
The answer and the passage are taken as the model input, while the question is the target output.
ROUGE-L~\cite{DBLP:conf/naacl/LinH03}, BLEU, and METEOR \cite{DBLP:conf/acl/BanerjeeL05} are treated as the assessment metrics.

\vspace{-0.3em}
\paragraph{Results}
As illustrated in Table~\ref{table_mt}, the randomly initialized NMT model yields comparable results with the reported system with the same architecture~\citep[26.1 vs. 26.0,][]{DBLP:journals/tacl/RotheNS20}, making our subsequent experiments convincing. Our methods significantly boost NLG performances across different pre-trained models, downstream tasks, linguistic resources, as well as segmentation tools, demonstrating its universal-effectiveness. Moreover, the embedding generator is able to decrease the vocabulary size and the generated sentence length, leading to less computational costs.


\section{Analysis}
To better understand subword discrepancy and our method, we make in-depth analyses on WMT En$\Rightarrow$De task to investigate three problems: 
\textbf{Q1}: How subword granularity affects NLG models? (\S~\ref{sec:5.1}) \textbf{Q2}: How embedding transfer benefits to downstream models? (\S~\ref{sec:5.2}) \textbf{Q3}: Dose our strategy acquire large computational costs? (\S~\ref{sec:5.4}) \textbf{Q4}: Can our
strategy exactly handle under-represented tokens? (\S~\ref{sec:5.3})
\begin{figure}[t] 
\centering 
\includegraphics[width=0.4\textwidth]{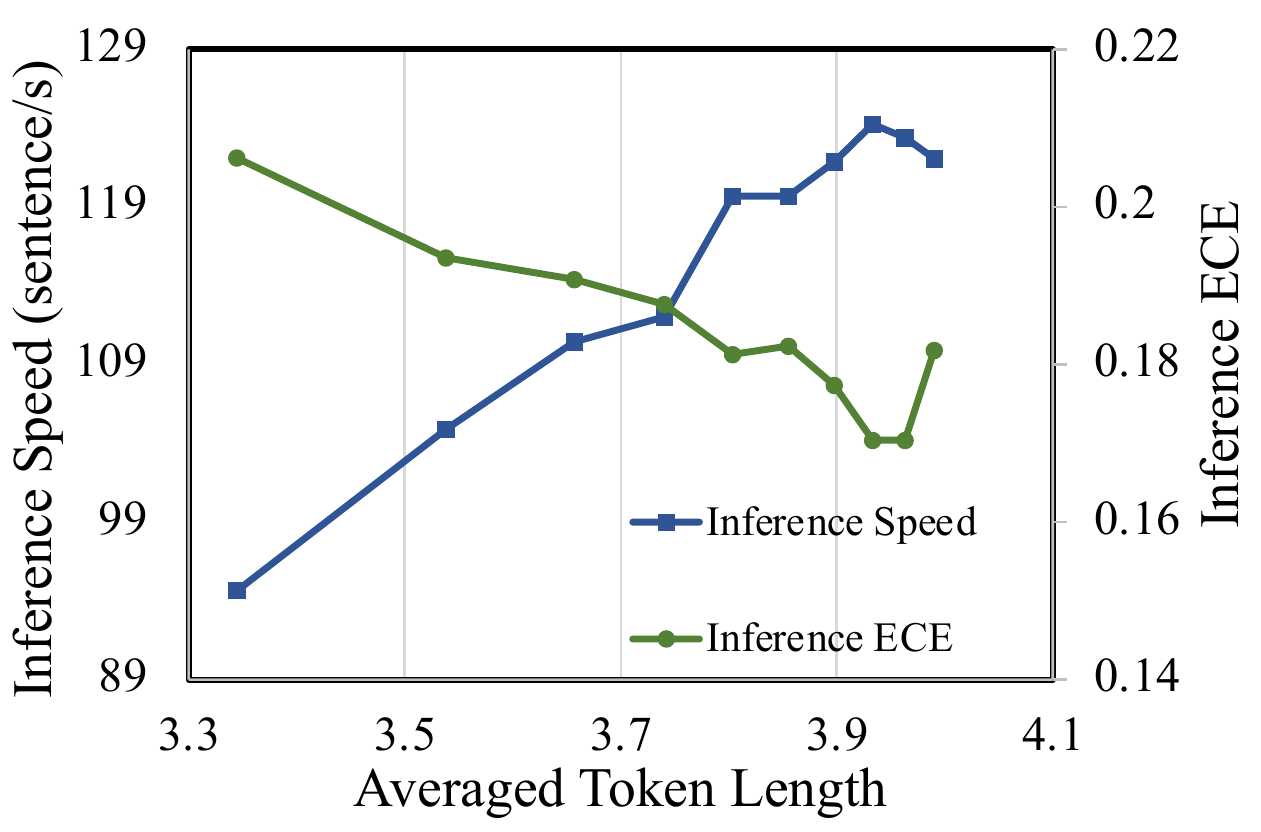} 
\caption{Effects of different token granularities on En$\Rightarrow$De task. As seen, the segmentation granularity remarkably affects inference speed and inference ECE.} 
\label{bpe_bleu}
\end{figure}

\subsection{Impact of Subword Granularity} \label{sec:5.1}
Figure \ref{bpe_bleu} visualizes the inference speed and exposure bias ~\citep[Inference Expected Calibration Error (ECE),][]{DBLP:conf/acl/WangTSL20} of translation models with different token granularities in their vocabulary. Obviously, for a translation model, neither too small nor too large granularity regarding to subwords can reach a satisfactory performance on inference speed. At the same time, the granularity indeed affects the problem of exposure bias in translation task. The experiments confirm the suitable segmentation strategy can effectively alleviate the problem of exposure bias.


\subsection{Impact of Embedding Transfer} \label{sec:5.2}
\begin{figure}[t] 
\centering 
\includegraphics[width=0.36\textwidth]{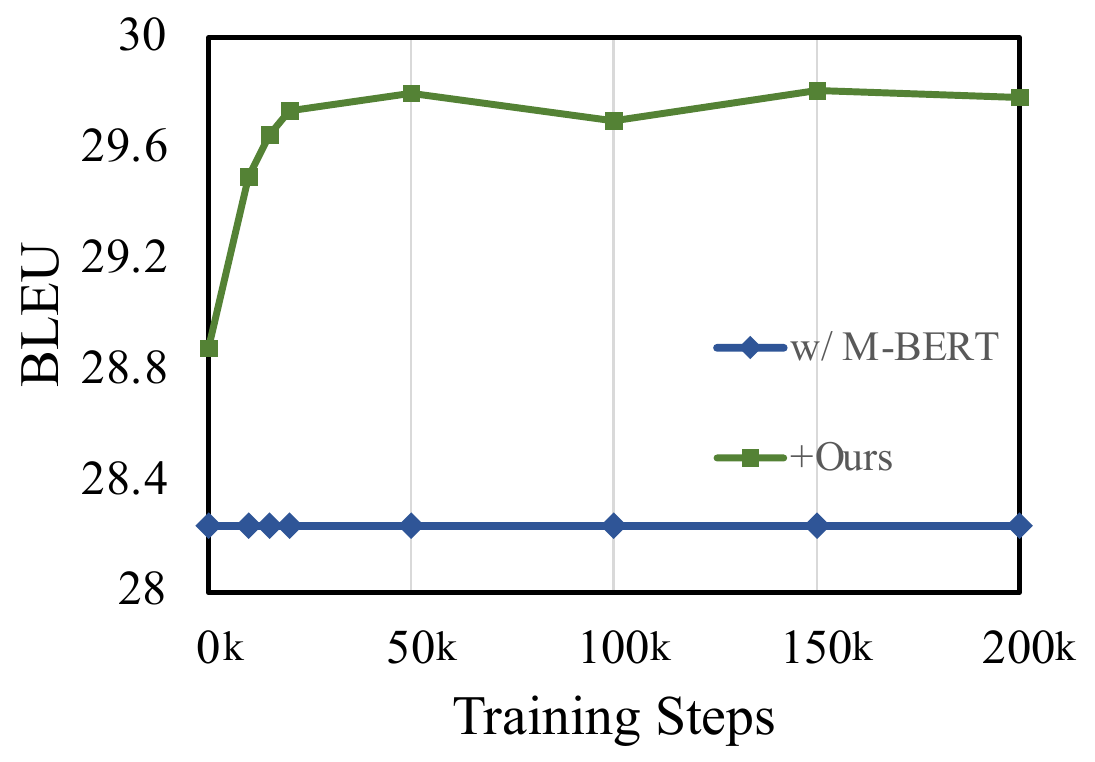} 
\caption{Effects of the training steps of embedding generators on BLEU scores of downstream models.} 
\label{loss&train}
\end{figure}

We further investigate how the embedding transfer impacts the initialization of downstream models. 
We draw Figure \ref{loss&train} to plot the BLEU scores of downstream models using the embedding generators trained with different steps. The X-axis indicates the training steps of the generator. Both “+Ours” and “w/ M-BERT” are fully finetuned, but the latter doesn't employ our embedding generator, resulting in an unchanged line. It is encouraging to see that the BLEU scores of downstream model converges very fast, indicating that our generator can be used with only a few of training steps. We argue that the commonalities in word compositionality lead to the fast transfer learning on generating different embeddings, and the simple architecture of our generator further speeds up such procedure.

\begin{table}[]
\scalebox{0.85}{
\begin{tabular}{l|ll}
\hline \hline
\multirow{2}{*}{\begin{tabular}[c]{@{}c@{}}Segmented\\ Token\end{tabular}} & \multicolumn{2}{l}{M-BERT: dan\textcolor{red}{\_}k\textcolor{red}{\_}bar}                                                                                                                                   \\ \cline{2-3} 
                                                                          & \multicolumn{2}{l}{Ours: dankbar}                                                                                                                                         \\ \hline \hline
\multirow{1}{*}{Source}                                                  & \multicolumn{2}{l}{\begin{tabular}[c]{@{}l@{}}it's very gratifying to have this kind \\ of reception here.\end{tabular}}         \\ \cline{1-3} 
                                    \multirow{1}{*}{Reference}                                       & \multicolumn{2}{l}{\begin{tabular}[c]{@{}l@{}} ich bin sehr \textcolor{red}{dankbar} für den \\ empfang hier.\end{tabular}} \\ \hline
\multirow{3}{*}{Translations}                                               & \multicolumn{2}{l}{\begin{tabular}[c]{@{}l@{}}M-BERT: es ist sehr befriedigend, \\ diese art von empfang hier zu haben.\end{tabular}}             \\ \cline{2-3} 
                                                                          & \multicolumn{2}{l}{\begin{tabular}[c]{@{}l@{}}Ours: ich bin sehr \textcolor{red}{dankbar} für den \\ empfang hier.\end{tabular}}             \\ \hline \hline
\end{tabular}
}
\caption{The German word \emph{dankbar} (gratifying) is over segmented by M-BERT, and mistranslated by its associated translation model. Our method can exactly approach this problem via using a more suitable segmentation for downstream tasks.  }
\label{case1}
\end{table}

\subsection{Computational Costs} \label{sec:5.4}
As shown in Figure \ref{loss&train}, our generator converges very fast (around 20K steps). The training process of our generator takes about 2 hours under our experimental setting. As a reference, the vanilla WMT finetuning process takes approximately 40 hours. In addition, our generator only takes about 3 minutes for producing 13K embeddings in Thesis, which is also insignificant compare to the finetuning time. Most importantly, once the embedding generator is well-trained, it's available for any downstream tasks. Thus, we argue that the computational costs are not the obstacle to the extensibility of our approach.

\subsection{Qualitative Analysis} \label{sec:5.3}
Table~\ref{case1} gives an example to show the effectiveness of our model on handling under-represented tokens. The German word \emph{dankbar} (gratifying) is over segmented by M-BERT, and fail to be generated by the model trained under conventional pipeline. On the contrary, our approach offers an opportunity for the downstream model to preserve the word into vocabulary, thus better learning its semantic meaning and correctly predicting it during inference.   


\section{Conclusion}

In this paper, we point out that the one-size-fits-all subword vocabulary, despite its all-encompassing superiority, is not the preferred solution for the popular pretrain-finetune paradigm. It causes the subword discrepancy among upstream and downstream models, which is given concrete form to the unsuitable granularity and under-represented words. Consequently, we propose a novel embedding transfer strategy with a plug-and-play embedding generator. Empirical results suggest that: 1) our approach is universally effective on overcoming subword discrepancy; 2) embedding transfer can bring benefits to computational efficiency; and 3) embedding generator can be achieved via either directly averaging the input embeddings or applying trainable components, the latter performs better but depends on few of training. 
As our approach is transparent to model architectures and tasks, we believe it can be widely applied and further raise the flexibility and applicability of pre-trained models. 

In the future, we plan to investigate its effectiveness on other generation tasks, such as code generation \cite{DBLP:conf/acl/huijiang21, DBLP:conf/acl/bbx21}, summarization \cite{DBLP:journals/tdasci/ShiKRR21} and so on.




\section*{Acknowledgments}
The project was supported by National Natural Science Foundation of China (No. 62036004, No. 61672440), National Key Research and Development Program of China (No. 2018YFB1403202),
Natural Science Foundation of Fujian Province of China (No. 2020J06001),
Youth Innovation Fund of Xiamen (No. 3502Z20206059),
and the Fundamental Research Funds for the Central Universities (No. ZK20720200077).
We also thank the reviewers for their insightful comments.

\bibliographystyle{acl_natbib}
\bibliography{anthology,acl2021}

\begin{thebibliography}{47}
\expandafter\ifx\csname natexlab\endcsname\relax\def\natexlab#1{#1}\fi

\bibitem[{Aji et~al.(2020)Aji, Bogoychev, Heafield, and
  Sennrich}]{DBLP:conf/acl/AjiBHS20}
Alham~Fikri Aji, Nikolay Bogoychev, Kenneth Heafield, and Rico Sennrich. 2020.
\newblock \href {https://www.aclweb.org/anthology/2020.acl-main.688/} {In
  neural machine translation, what does transfer learning transfer?}
\newblock In \emph{ACL 2020}.

\bibitem[{Bahdanau et~al.(2015)Bahdanau, Cho, and
  Bengio}]{DBLP:journals/corr/BahdanauCB14}
Dzmitry Bahdanau, Kyunghyun Cho, and Yoshua Bengio. 2015.
\newblock \href {http://arxiv.org/abs/1409.0473} {Neural machine translation by
  jointly learning to align and translate}.
\newblock In \emph{ICLR 2015}.

\bibitem[{Banerjee and Lavie(2005)}]{DBLP:conf/acl/BanerjeeL05}
Satanjeev Banerjee and Alon Lavie. 2005.
\newblock \href {https://www.aclweb.org/anthology/W05-0909/} {{METEOR}: An
  automatic metric for {MT} evaluation with improved correlation with human
  judgments}.
\newblock In \emph{ACL 2005}.

\bibitem[{Bengio et~al.(2015)Bengio, Vinyals, Jaitly, and
  Shazeer}]{DBLP:conf/nips/BengioVJS15}
Samy Bengio, Oriol Vinyals, Navdeep Jaitly, and Noam Shazeer. 2015.
\newblock \href
  {http://papers.nips.cc/paper/5956-scheduled-sampling-for-sequence-prediction-with-recurrent-neural-networks}
  {Scheduled sampling for sequence prediction with recurrent neural networks}.
\newblock In \emph{NIPS 2015}.

\bibitem[{Cherry et~al.(2018)Cherry, Foster, Bapna, Firat, and
  Macherey}]{DBLP:conf/emnlp/CherryFBFM18}
Colin Cherry, George Foster, Ankur Bapna, Orhan Firat, and Wolfgang Macherey.
  2018.
\newblock \href {https://www.aclweb.org/anthology/D18-1461/} {Revisiting
  character-based neural machine translation with capacity and compression}.
\newblock In \emph{EMNLP 2018}.

\bibitem[{Chronopoulou et~al.(2019)Chronopoulou, Baziotis, and
  Potamianos}]{DBLP:conf/naacl/ChronopoulouBP19}
Alexandra Chronopoulou, Christos Baziotis, and Alexandros Potamianos. 2019.
\newblock \href {https://doi.org/10.18653/v1/n19-1213} {An embarrassingly
  simple approach for transfer learning from pretrained language models}.
\newblock In \emph{NAACL 2019}.

\bibitem[{Devlin et~al.(2019)Devlin, Chang, Lee, and
  Toutanova}]{DBLP:conf/naacl/DevlinCLT19}
Jacob Devlin, Ming-Wei Chang, Kenton Lee, and Kristina Toutanova. 2019.
\newblock \href {https://doi.org/10.18653/v1/n19-1423} {{BERT}: Pre-training of
  deep bidirectional transformers for language understanding}.
\newblock In \emph{ACL 2019}.

\bibitem[{Edunov et~al.(2019)Edunov, Baevski, and
  Auli}]{DBLP:conf/naacl/EdunovBA19}
Sergey Edunov, Alexei Baevski, and Michael Auli. 2019.
\newblock \href {https://doi.org/10.18653/v1/n19-1409} {Pre-trained language
  model representations for language generation}.
\newblock In \emph{ACL 2019}.

\bibitem[{Fukuda et~al.(2020)Fukuda, Yoshinaga, and
  Kitsuregawa}]{DBLP:conf/emnlp/Fukuda0K20}
Nobukazu Fukuda, Naoki Yoshinaga, and Masaru Kitsuregawa. 2020.
\newblock \href {https://doi.org/10.18653/v1/2020.findings-emnlp.434} {Robust
  {B}acked-off {E}stimation of {O}ut-of-{V}ocabulary {E}mbeddings}.
\newblock In \emph{EMNLP Findings 2020}.

\bibitem[{Hinton et~al.(2015)Hinton, Vinyals, and
  Dean}]{DBLP:journals/corr/HintonVD15}
Geoffrey~E. Hinton, Oriol Vinyals, and Jeffrey Dean. 2015.
\newblock \href {http://arxiv.org/abs/1503.02531} {Distilling the knowledge in
  a neural network}.
\newblock \emph{CoRR 2015}, abs/1503.02531.

\bibitem[{Jiang et~al.(2021)Jiang, Zhou, Meng, Zhang, Zhou, Huang, Wu, and
  Su}]{DBLP:conf/acl/huijiang21}
Hui Jiang, Chulun Zhou, Fandong Meng, Biao Zhang, Jie Zhou, Degen Huang,
  Qingqiang Wu, and Jinsong Su. 2021.
\newblock Exploring dynamic selection of branch expansion orders for code
  generation.
\newblock In \emph{ACL 2021}.

\bibitem[{Khayrallah et~al.(2018)Khayrallah, Thompson, Duh, and
  Koehn}]{DBLP:conf/aclnmt/KhayrallahTDK18}
Huda Khayrallah, Brian Thompson, Kevin Duh, and Philipp Koehn. 2018.
\newblock \href {https://doi.org/10.18653/v1/w18-2705} {Regularized training
  objective for continued training for domain adaptation in neural machine
  translation}.
\newblock In \emph{Proceedings of the 2nd Workshop on Neural Machine
  Translation and Generation 2018}.

\bibitem[{Kingma and Ba(2015)}]{DBLP:journals/corr/KingmaB14}
Diederik~P. Kingma and Jimmy Ba. 2015.
\newblock \href {http://arxiv.org/abs/1412.6980} {Adam: {A} method for
  stochastic optimization}.
\newblock In \emph{ICLR 2015}.

\bibitem[{Kocmi and Bojar(2018)}]{DBLP:conf/wmt/KocmiB18}
Tom Kocmi and Ond{\v{r}}ej Bojar. 2018.
\newblock \href {https://doi.org/10.18653/v1/w18-6325} {Trivial transfer
  learning for low-resource neural machine translation}.
\newblock In \emph{Machine Translation: Research Papers 2018}.

\bibitem[{Kudo(2018)}]{DBLP:conf/acl/Kudo18}
Taku Kudo. 2018.
\newblock \href {https://www.aclweb.org/anthology/P18-1007/} {Subword
  regularization: Improving neural network translation models with multiple
  subword candidates}.
\newblock In \emph{ACL 2018}.

\bibitem[{Kudo and Richardson(2018)}]{DBLP:conf/emnlp/KudoR18}
Taku Kudo and John Richardson. 2018.
\newblock \href {https://doi.org/10.18653/v1/d18-2012} {{S}entence{P}iece: A
  simple and language independent subword tokenizer and detokenizer for neural
  text processing}.
\newblock In \emph{EMNLP 2018}.

\bibitem[{Lewis et~al.(2020)Lewis, Liu, Goyal, Ghazvininejad, Mohamed, Levy,
  Stoyanov, and Zettlemoyer}]{DBLP:conf/acl/LewisLGGMLSZ20}
Mike Lewis, Yinhan Liu, Naman Goyal, Marjan Ghazvininejad, Abdelrahman Mohamed,
  Omer Levy, Veselin Stoyanov, and Luke Zettlemoyer. 2020.
\newblock \href {https://www.aclweb.org/anthology/2020.acl-main.703/} {{BART}:
  Denoising sequence-to-sequence pre-training for natural language generation,
  translation, and comprehension}.
\newblock In \emph{ACL 2020}.

\bibitem[{Lin and Hovy(2003)}]{DBLP:conf/naacl/LinH03}
Chin-Yew Lin and Eduard Hovy. 2003.
\newblock \href {https://www.aclweb.org/anthology/N03-1020/} {Automatic
  evaluation of summaries using n-gram co-occurrence statistics}.
\newblock In \emph{NAACL 2003}.

\bibitem[{Liu et~al.(2020{\natexlab{a}})Liu, Yan, Gong, Qi, Zhang, Jiao, Chen,
  Fu, Shou, Gong, Wang, Chen, Jiang, Lv, Zhang, Wu, Zhou, and
  Duan}]{DBLP:journals/corr/abs-2011-11928}
Dayiheng Liu, Yu~Yan, Yeyun Gong, Weizhen Qi, Hang Zhang, Jian Jiao, Weizhu
  Chen, Jie Fu, Linjun Shou, Ming Gong, Pengcheng Wang, Jiusheng Chen, Daxin
  Jiang, Jiancheng Lv, Ruofei Zhang, Winnie Wu, Ming Zhou, and Nan Duan.
  2020{\natexlab{a}}.
\newblock \href {https://arxiv.org/abs/2011.11928} {{GLGE:} {A} new general
  language generation evaluation benchmark}.
\newblock \emph{CoRR 2020}, abs/2011.11928.

\bibitem[{Liu et~al.(2020{\natexlab{b}})Liu, Liu, Li, Su, Ge, Wang, and
  Luo}]{DBLP:conf/ijcai/LiuLLSGWL20}
Xin Liu, Kai Liu, Xiang Li, Jinsong Su, Yubin Ge, Bin Wang, and Jiebo Luo.
  2020{\natexlab{b}}.
\newblock \href {https://doi.org/10.24963/ijcai.2020/525} {An iterative
  multi-source mutual knowledge transfer framework for machine reading
  comprehension}.
\newblock In \emph{IJCAI 2020}.

\bibitem[{Liu et~al.(2020{\natexlab{c}})Liu, Gu, Goyal, Li, Edunov,
  Ghazvininejad, Lewis, and Zettlemoyer}]{liu2020multilingual}
Yinhan Liu, Jiatao Gu, Naman Goyal, Xian Li, Sergey Edunov, Marjan
  Ghazvininejad, Mike Lewis, and Luke Zettlemoyer. 2020{\natexlab{c}}.
\newblock \href {https://arxiv.org/abs/2001.08210} {Multilingual denoising
  pre-training for neural machine translation}.
\newblock \emph{TACL 2020}.

\bibitem[{Mikolov et~al.(2013)Mikolov, Chen, Corrado, and
  Dean}]{DBLP:journals/corr/abs-1301-3781}
Tom{\'{a}}s Mikolov, Kai Chen, Greg Corrado, and Jeffrey Dean. 2013.
\newblock \href {http://arxiv.org/abs/1301.3781} {Efficient estimation of word
  representations in vector space}.
\newblock In \emph{ICLR 2013}.

\bibitem[{Nguyen and Chiang(2017)}]{DBLP:conf/ijcnlp/NguyenC17}
Toan~Q. Nguyen and David Chiang. 2017.
\newblock \href {https://www.aclweb.org/anthology/I17-2050/} {Transfer learning
  across low-resource, related languages for neural machine translation}.
\newblock In \emph{IJCNLP 2017}.

\bibitem[{Papineni et~al.(2002)Papineni, Roukos, Ward, and
  Zhu}]{DBLP:conf/acl/PapineniRWZ02}
Kishore Papineni, Salim Roukos, Todd Ward, and Wei-Jing Zhu. 2002.
\newblock \href {https://www.aclweb.org/anthology/P02-1040/} {{B}leu: a method
  for automatic evaluation of machine translation}.
\newblock In \emph{ACL 2002}.

\bibitem[{Pinter et~al.(2017)Pinter, Guthrie, and
  Eisenstein}]{DBLP:conf/emnlp/PinterGE17}
Yuval Pinter, Robert Guthrie, and Jacob Eisenstein. 2017.
\newblock \href {https://doi.org/10.18653/v1/d17-1010} {Mimicking word
  embeddings using subword {RNN}s}.
\newblock In \emph{EMNLP 2017}.

\bibitem[{Provilkov et~al.(2020)Provilkov, Emelianenko, and
  Voita}]{DBLP:conf/acl/ProvilkovEV20}
Ivan Provilkov, Dmitrii Emelianenko, and Elena Voita. 2020.
\newblock \href {https://doi.org/10.18653/v1/2020.acl-main.170} {{BPE}-dropout:
  Simple and effective subword regularization}.
\newblock In \emph{ACL 2020}.

\bibitem[{Radford et~al.(2019)Radford, Wu, Child, Luan, Amodei, and
  Sutskever}]{radford2019language}
Alec Radford, Jeffrey Wu, Rewon Child, David Luan, Dario Amodei, and Ilya
  Sutskever. 2019.
\newblock \href {http://www.persagen.com/files/misc/radford2019language.pdf}
  {Language models are unsupervised multitask learners}.
\newblock \emph{OpenAI blog 2019}, 1(8):9.

\bibitem[{Rajpurkar et~al.(2016)Rajpurkar, Zhang, Lopyrev, and
  Liang}]{DBLP:conf/emnlp/RajpurkarZLL16}
Pranav Rajpurkar, Jian Zhang, Konstantin Lopyrev, and Percy Liang. 2016.
\newblock \href {https://doi.org/10.18653/v1/d16-1264} {{SQ}u{AD}: 100,000+
  questions for machine comprehension of text}.
\newblock In \emph{EMNLP 2016}.

\bibitem[{Rothe et~al.(2020)Rothe, Narayan, and
  Severyn}]{DBLP:journals/tacl/RotheNS20}
Sascha Rothe, Shashi Narayan, and Aliaksei Severyn. 2020.
\newblock \href {https://transacl.org/ojs/index.php/tacl/article/view/1849}
  {Leveraging pre-trained checkpoints for sequence generation tasks}.
\newblock \emph{TACL 2020}.

\bibitem[{Sasaki et~al.(2019)Sasaki, Suzuki, and
  Inui}]{DBLP:conf/naacl/SasakiSI19}
Shota Sasaki, Jun Suzuki, and Kentaro Inui. 2019.
\newblock \href {https://doi.org/10.18653/v1/n19-1353} {{S}ubword-based
  {C}ompact {R}econstruction of {W}ord {E}mbeddings}.
\newblock In \emph{NAACL 2019}.

\bibitem[{Sato et~al.(2020)Sato, Sakuma, Yoshinaga, Toyoda, and
  Kitsuregawa}]{DBLP:conf/emnlp/SatoS0TK20}
Shoetsu Sato, Jin Sakuma, Naoki Yoshinaga, Masashi Toyoda, and Masaru
  Kitsuregawa. 2020.
\newblock \href {https://doi.org/10.18653/v1/2020.findings-emnlp.381}
  {Vocabulary adaptation for domain adaptation in neural machine translation}.
\newblock In \emph{EMNLP 2020}.

\bibitem[{Sennrich et~al.(2016)Sennrich, Haddow, and
  Birch}]{DBLP:conf/acl/SennrichHB16a}
Rico Sennrich, Barry Haddow, and Alexandra Birch. 2016.
\newblock \href {https://doi.org/10.18653/v1/p16-1162} {Neural machine
  translation of rare words with subword units}.
\newblock In \emph{ACL 2016}.

\bibitem[{Shi et~al.(2021)Shi, Keneshloo, Ramakrishnan, and
  Reddy}]{DBLP:journals/tdasci/ShiKRR21}
Tian Shi, Yaser Keneshloo, Naren Ramakrishnan, and Chandan~K. Reddy. 2021.
\newblock \href {https://doi.org/10.1145/3419106} {Neural abstractive text
  summarization with sequence-to-sequence models}.
\newblock \emph{Trans. Data Sci.}, 2(1):1:1--1:37.

\bibitem[{Su et~al.(2021)Su, Zeng, Xie, Wen, Yin, and Liu}]{jiali_tpami}
Jinsong Su, Jiali Zeng, Jun Xie, Huating Wen, Yongjing Yin, and Yang Liu. 2021.
\newblock \href {https://doi.org/10.1109/TPAMI.2019.2954406} {Exploring
  discriminative word-level domain contexts for multi-domain neural machine
  translation}.
\newblock \emph{IEEE Transactions on Pattern Analysis and Machine
  Intelligence}, 43(5):1530--1545.

\bibitem[{Tian et~al.(2014)Tian, Wong, Chao, Quaresma, Oliveira, Lu, Li, Wang,
  and Wang}]{DBLP:conf/lrec/TianWCQOY14}
Liang Tian, Derek~F. Wong, Lidia~S. Chao, Paulo Quaresma, Francisco Oliveira,
  Yi~Lu, Shuo Li, Yiming Wang, and Longyue Wang. 2014.
\newblock \href
  {http://www.lrec-conf.org/proceedings/lrec2014/summaries/774.html}
  {{UM}-corpus: A large {E}nglish-{C}hinese parallel corpus for statistical
  machine translation}.
\newblock In \emph{LREC 2014}.

\bibitem[{Vaswani et~al.(2017)Vaswani, Shazeer, Parmar, Uszkoreit, Jones,
  Gomez, Kaiser, and Polosukhin}]{DBLP:conf/nips/VaswaniSPUJGKP17}
Ashish Vaswani, Noam Shazeer, Niki Parmar, Jakob Uszkoreit, Llion Jones,
  Aidan~N. Gomez, Lukasz Kaiser, and Illia Polosukhin. 2017.
\newblock \href
  {https://proceedings.neurips.cc/paper/2017/hash/3f5ee243547dee91fbd053c1c4a845aa-Abstract.html}
  {Attention is all you need}.
\newblock In \emph{NIPS 2017}.

\bibitem[{Wan et~al.(2020)Wan, Yang, Wong, Zhou, Chao, Zhang, and
  Chen}]{DBLP:conf/emnlp/WanYWZCZC20}
Yu~Wan, Baosong Yang, Derek~F. Wong, Yikai Zhou, Lidia~S. Chao, Haibo Zhang,
  and Boxing Chen. 2020.
\newblock \href {https://doi.org/10.18653/v1/2020.emnlp-main.80} {Self-paced
  learning for neural machine translation}.
\newblock In \emph{EMNLP 2020}, pages 1074--1080.

\bibitem[{Wang et~al.(2020)Wang, Tu, Shi, and Liu}]{DBLP:conf/acl/WangTSL20}
Shuo Wang, Zhaopeng Tu, Shuming Shi, and Yang Liu. 2020.
\newblock \href {https://doi.org/10.18653/v1/2020.acl-main.278} {On the
  inference calibration of neural machine translation}.
\newblock In \emph{ACL 2020}.

\bibitem[{Wu et~al.(2016)Wu, Schuster, Chen, Le, Norouzi, Macherey, Krikun,
  Cao, Gao, Macherey, Klingner, Shah, Johnson, Liu, Kaiser, Gouws, Kato, Kudo,
  Kazawa, Stevens, Kurian, Patil, Wang, Young, Smith, Riesa, Rudnick, Vinyals,
  Corrado, Hughes, and Dean}]{DBLP:journals/corr/WuSCLNMKCGMKSJL16}
Yonghui Wu, Mike Schuster, Zhifeng Chen, Quoc~V. Le, Mohammad Norouzi, Wolfgang
  Macherey, Maxim Krikun, Yuan Cao, Qin Gao, Klaus Macherey, Jeff Klingner,
  Apurva Shah, Melvin Johnson, Xiaobing Liu, Lukasz Kaiser, Stephan Gouws,
  Yoshikiyo Kato, Taku Kudo, Hideto Kazawa, Keith Stevens, George Kurian,
  Nishant Patil, Wei Wang, Cliff Young, Jason Smith, Jason Riesa, Alex Rudnick,
  Oriol Vinyals, Greg Corrado, Macduff Hughes, and Jeffrey Dean. 2016.
\newblock \href {http://arxiv.org/abs/1609.08144} {Google's neural machine
  translation system: Bridging the gap between human and machine translation}.
\newblock \emph{CoRR 2016}, abs/1609.08144.

\bibitem[{Xie et~al.(2021)Xie, Su, Li, Ge, Cui, Yao, and
  and}]{DBLP:conf/acl/bbx21}
Binbin Xie, Jinsong Su, Xiang Li, Yubin Ge, Jianwei Cui, Junfeng Yao, and
  Bin~Wang and. 2021.
\newblock Improving tree-structured decoder training for code generation via
  mutual learning.
\newblock In \emph{AAAI 2021}.

\bibitem[{Xu et~al.(2019)Xu, Liu, Shu, and Yu}]{DBLP:conf/naacl/XuLSY19}
Hu~Xu, Bing Liu, Lei Shu, and Philip Yu. 2019.
\newblock \href {https://doi.org/10.18653/v1/n19-1242} {{BERT} post-training
  for review reading comprehension and aspect-based sentiment analysis}.
\newblock In \emph{ACL 2019}.

\bibitem[{Yang et~al.(2019{\natexlab{a}})Yang, Wang, Liu, Liu, Lyu, Wu, She,
  and Li}]{DBLP:conf/acl/YangWLLLWSL19}
An~Yang, Quan Wang, Jing Liu, Kai Liu, Yajuan Lyu, Hua Wu, Qiaoqiao She, and
  Sujian Li. 2019{\natexlab{a}}.
\newblock \href {https://doi.org/10.18653/v1/p19-1226} {Enhancing pre-trained
  language representations with rich knowledge for machine reading
  comprehension}.
\newblock In \emph{ACL 2019}.

\bibitem[{Yang et~al.(2019{\natexlab{b}})Yang, Dai, Yang, Carbonell,
  Salakhutdinov, and Le}]{DBLP:conf/nips/YangDYCSL19}
Zhilin Yang, Zihang Dai, Yiming Yang, Jaime~G. Carbonell, Ruslan Salakhutdinov,
  and Quoc~V. Le. 2019{\natexlab{b}}.
\newblock \href
  {http://papers.nips.cc/paper/8812-xlnet-generalized-autoregressive-pretraining-for-language-understanding}
  {Xlnet: Generalized autoregressive pretraining for language understanding}.
\newblock In \emph{NeurIPS 2019}.

\bibitem[{Yao et~al.(2020)Yao, Yang, Zhang, Chen, and Luo}]{coling/YaoYZCL20}
Liang Yao, Baosong Yang, Haibo Zhang, Boxing Chen, and Weihua Luo. 2020.
\newblock \href {https://doi.org/10.18653/v1/2020.coling-main.399} {Domain
  transfer based data augmentation for neural query translation}.
\newblock In \emph{COLING 2020}.

\bibitem[{Zeng et~al.(2018)Zeng, Su, Wen, Liu, Xie, Yin, and
  Zhao}]{DBLP:conf/emnlp/ZengSWLXYZ18}
Jiali Zeng, Jinsong Su, Huating Wen, Yang Liu, Jun Xie, Yongjing Yin, and
  Jianqiang Zhao. 2018.
\newblock \href {https://doi.org/10.18653/v1/d18-1041} {Multi-domain neural
  machine translation with word-level domain context discrimination}.
\newblock In \emph{EMNLP 2018}.

\bibitem[{Zhao et~al.(2018)Zhao, Mudgal, and Liang}]{DBLP:conf/emnlp/ZhaoML18}
Jinman Zhao, Sidharth Mudgal, and Yingyu Liang. 2018.
\newblock \href {https://doi.org/10.18653/v1/d18-1059} {Generalizing word
  embeddings using bag of subwords}.
\newblock In \emph{EMNLP 2018}.

\bibitem[{Zhu et~al.(2020)Zhu, Xia, Wu, He, Qin, Zhou, Li, and
  Liu}]{DBLP:conf/iclr/ZhuXWHQZLL20}
Jinhua Zhu, Yingce Xia, Lijun Wu, Di~He, Tao Qin, Wengang Zhou, Houqiang Li,
  and Tie{-}Yan Liu. 2020.
\newblock \href {https://openreview.net/forum?id=Hyl7ygStwB} {Incorporating
  {BERT} into neural machine translation}.
\newblock In \emph{ICLR 2020}.

\end{thebibliography}


\end{document}